\documentclass[11pt]{article}

\usepackage[utf8]{inputenc}
\usepackage[T1]{fontenc}
\usepackage{times}
\usepackage[margin=1in]{geometry}
\usepackage{amsmath,amssymb}
\usepackage{graphicx}
\usepackage{booktabs}
\usepackage{hyperref}
\usepackage{xcolor}
\usepackage{natbib}
\usepackage{caption}
\usepackage{subcaption}
\usepackage{enumitem}
\usepackage{multirow}
\usepackage{array}

\hypersetup{
  colorlinks=true,
  linkcolor=blue!60!black,
  citecolor=blue!60!black,
  urlcolor=blue!60!black
}

\title{\Large\textbf{The Yerkes-Dodson Curve for AI Agents:\\Optimal Environmental Pressure for Emergent Complexity\\in LLM Multi-Agent Systems}}

\author{
  Ivan Pasichnyk\\
  WeLabelData Inc.\\
  \texttt{ivan@welabeldata.com}\\
  ORCID: \href{https://orcid.org/0009-0004-8154-3345}{0009-0004-8154-3345}
}

\date{March 2026}

\begin{document}
\maketitle

\begin{abstract}
Designing environments that maximize the rate of emergent behavior development in AI agents remains an open problem.
We present the first systematic study of stress-performance relationships in large language model (LLM) multi-agent systems, drawing an explicit parallel to the Yerkes-Dodson law from cognitive psychology.
Using a grid-world survival arena, we conduct 22 experiments across four phases, varying environmental pressure through resource scarcity (upkeep cost) and reproductive competition (sexual selection).
Our key finding is that cooperative behavior follows an inverted-U curve: trade interactions peak at 29 under medium pressure (upkeep=5), while both low and extreme pressure produce 8--12 trades.
Under extreme pressure, behavioral repertoire collapses to movement-only within 5--12 turns.
We further show that sexual selection---a softer pressure mechanism where all agents survive but not all reproduce---eliminates inter-agent aggression entirely and produces communicative behavior absent under survival pressure.
These results suggest that environmental pressure calibration is a viable curriculum design strategy for LLM agent development, analogous to the inverted-U relationship between arousal and performance in biological systems.
\end{abstract}

\section{Introduction}

The capability frontier of AI systems increasingly depends not just on model scale or training data, but on the environments in which agents operate and develop~\citep{baker2020emergent, team2021openended}.
As large language models (LLMs) are deployed as autonomous agents in complex, multi-agent settings~\citep{park2023generative}, a critical design question emerges: \emph{how should we calibrate environmental difficulty to maximize the emergence of complex social behavior?}

This question has a well-studied analog in cognitive psychology.
The Yerkes-Dodson law~\citep{yerkes1908relation, diamond2007yerkes} describes an inverted-U relationship between arousal (stress) and task performance: too little arousal leads to disengagement, too much leads to anxiety and performance collapse, and optimal performance occurs at intermediate levels.
Despite its foundational status in psychology, this principle has never been systematically tested in LLM agent populations.

Recent work has established that different LLMs exhibit distinct, stable behavioral profiles in game-theoretic settings~\citep{fan2025understanding, chen2025strategic, wang2025reasoning}---effectively acting as different ``species'' with characteristic cooperation, defection, and aggression tendencies.
Studies have also demonstrated that LLM agents exhibit survival instincts under resource scarcity~\citep{xie2025survival} and that personality traits emerge consistently across models~\citep{huang2025personality, li2025personality}.
However, no prior work has systematically varied environmental pressure to map the stress-performance curve for LLM agent populations.

We address this gap with three research questions:

\begin{itemize}[leftmargin=2em,itemsep=2pt]
  \item[\textbf{RQ1}:] Do LLM agents exhibit a Yerkes-Dodson (inverted-U) relationship between environmental pressure and cooperative behavior?
  \item[\textbf{RQ2}:] At what pressure level does behavioral repertoire collapse, and what does collapse look like?
  \item[\textbf{RQ3}:] Can reproductive competition---as opposed to survival threat---drive social complexity without lethal consequences?
\end{itemize}

To answer these questions, we introduce the \emph{Survival Arena}, a grid-world environment where LLM agents (Claude 3.5 Sonnet) make decisions about resource gathering, movement, combat, trade, and reproduction.
We systematically vary environmental pressure along two axes: survival cost (upkeep) and reproductive competition (sexual selection with provider/chooser dynamics).

Our contributions are:

\begin{enumerate}[leftmargin=2em,itemsep=2pt]
  \item We provide the first empirical demonstration of a Yerkes-Dodson curve in LLM multi-agent systems, showing that cooperative trade peaks at medium pressure and declines under both low and extreme pressure.
  \item We characterize behavioral collapse under extreme pressure: agents reduce to movement-only strategies within 5--12 turns, eliminating all social behavior.
  \item We introduce sexual selection as an alternative pressure mechanism and show it produces zero aggression and richer communication compared to survival pressure.
  \item We identify a methodological insight: Shannon entropy over whole-game action distributions is a misleading complexity metric due to small-sample confounds at high pressure levels.
\end{enumerate}

\section{Related Work}

\paragraph{LLM agents in game-theoretic settings.}
Several recent studies examine LLM behavior in strategic interactions.
\citet{fan2025understanding} find that Claude 3.5 exhibits the strongest prosocial bias among tested models, cooperating even under selfish framing.
\citet{chen2025strategic} show GPT-o1 excels in competitive/adversarial games but suffers paranoid trust collapse, while DeepSeek-R1 demonstrates superior cooperation, forward planning, and theory of mind.
\citet{wang2025reasoning} further document trust dynamics between DeepSeek and GPT variants.
These findings establish that LLMs have stable behavioral ``phenotypes,'' but do not explore how environmental pressure modulates these behaviors.

\paragraph{Survival instinct in LLM agents.}
\citet{xie2025survival} demonstrate that LLM agents exhibit survival instincts in a Sugarscape-style environment, with GPT-4o and Gemini showing attack rates exceeding 80\% under scarcity.
Their work examines survival behavior but does not systematically vary pressure levels to map the full stress-performance curve.

\paragraph{Multi-agent environments and emergent behavior.}
The study of emergent social behavior in agent populations has a long history, from Sugarscape~\citep{epstein1996growing} to recent work on emergent tool use from autocurricula~\citep{baker2020emergent}.
Self-play and asymmetric self-play~\citep{silver2017mastering, sukhbaatar2018intrinsic} have shown that competitive pressure drives capability development.
\citet{leibo2017multiagent} study cooperation and defection dynamics in sequential social dilemmas.
Our work extends this line by using LLMs as agent policies and varying environmental---rather than strategic---pressure.

\paragraph{Curriculum learning and open-ended evolution.}
The idea that training difficulty should follow a curriculum is well-established~\citep{team2021openended}.
However, most curriculum learning focuses on supervised or RL settings with gradient-based optimization.
Our approach treats LLM pretrained policies as fixed and varies the \emph{environment} rather than the model, creating a curriculum through evolutionary pressure rather than parameter updates.

\paragraph{Yerkes-Dodson law.}
The original Yerkes-Dodson law~\citep{yerkes1908relation} demonstrated an inverted-U relationship between stimulus intensity and habit formation in mice.
\citet{diamond2007yerkes} reviews its extension to human cognition, attention, and memory.
To our knowledge, we are the first to test this relationship in AI agent systems.

\paragraph{Sexual selection theory.}
Darwin's theory of sexual selection~\citep{darwin1871descent} was formalized by \citet{trivers1972parental} through parental investment theory and by \citet{zahavi1975mate} through the handicap principle.
\citet{byrne1988machiavellian} argue that social intelligence evolved primarily for mating competition.
We draw on these frameworks to design reproductive pressure mechanisms for LLM agents.

\section{Method: Survival Arena}

\subsection{Environment}

The Survival Arena is a discrete grid-world with resource nodes that regenerate each turn.
The v6.1 engine (used for survival pressure experiments) uses a $9 \times 9$ grid; the v7.0 engine (sexual selection) uses a $7 \times 7$ grid.
The environment contains two resource types:

\begin{itemize}[leftmargin=2em,itemsep=2pt]
  \item \textbf{Food nodes} ($n_f = 8$): Each regenerates $r_f$ food units per turn. Food is the primary survival resource; agents die when food reaches zero.
  \item \textbf{Token nodes} ($n_t = 5$): Each regenerates $r_t$ tokens per turn. Tokens serve as a secondary currency for trade and costly signaling.
\end{itemize}

Each grid cell has a maximum agent capacity of 3.
Agents start at random positions with initial resources proportional to their vitality level.

\subsection{Agent architecture}

Each agent is parameterized by six attributes drawn from $[1, 8]$ with a total budget of 30 points:

\begin{itemize}[leftmargin=2em,itemsep=2pt]
  \item \textbf{STR} (Strength): Determines attack damage and gathering efficiency.
  \item \textbf{SPD} (Speed): Agents with SPD $\geq 5$ can move 2 cells per turn.
  \item \textbf{INT} (Intelligence): Affects trade outcomes and training speed.
  \item \textbf{SOC} (Social): Controls observation accuracy of nearby agents and communication range.
  \item \textbf{END} (Endurance): Determines maximum health and rest recovery rate.
  \item \textbf{CHA} (Charisma): Influences trade acceptance probability.
\end{itemize}

Agent decisions are produced by an LLM (Claude 3.5 Sonnet) conditioned on a prompt containing: the agent's current state (resources, health, position, attributes), nearby agents' visible information, available actions, and recent action history.
Critically, the prompt contains no behavioral hints---agents receive no guidance on strategy, cooperation norms, or survival tips.

\subsection{Action space}

Agents choose one action per turn from:

\begin{itemize}[leftmargin=2em,itemsep=2pt]
  \item \textbf{GATHER}: Collect resources from a node at the agent's position.
  \item \textbf{MOVE}: Move to an adjacent cell (or 2 cells if SPD $\geq 5$).
  \item \textbf{ATTACK}: Deal STR-based damage to another agent at the same position. Costs 1 token.
  \item \textbf{TRADE}: Propose a resource exchange with a nearby agent.
  \item \textbf{REST}: Recover health proportional to END.
  \item \textbf{TRAIN}: Improve a chosen attribute (INT-dependent speed).
  \item \textbf{COMMUNICATE} (v7 only): Send a message to nearby agents. Costs 2 tokens; reveals full stats for 3 turns.
  \item \textbf{REPRODUCE} (v7 only): Propose reproduction with a nearby agent.
\end{itemize}

\subsection{Pressure axis 1: Upkeep (survival cost)}

Each turn, every agent pays an \emph{upkeep cost} $u$ in food.
If an agent's food drops to zero, it dies and is removed.
Higher upkeep creates survival pressure: agents must gather more efficiently, compete for scarce resources, or cooperate through trade to survive.
In Phase P2b, we hold all other parameters constant (food nodes, token nodes, regeneration rates) and vary only upkeep from 2 to 7.

\subsection{Pressure axis 2: Sexual selection (V7)}

In the V7 engine, we introduce reproductive competition based on Trivers' parental investment theory~\citep{trivers1972parental}.
Agents are assigned one of two types:

\begin{itemize}[leftmargin=2em,itemsep=2pt]
  \item \textbf{Providers} (8 agents): Can propose reproduction. Cost: 6 food + 3 tokens.
  \item \textbf{Choosers} (8 agents): Evaluate proposals via a separate LLM call. Cost: 12 food + 5 tokens.
\end{itemize}

Each agent has a \emph{vitality} level (frail/average/robust/radiant) affecting starting resources and offspring quality.
Offspring inherit averaged and mutated attributes from both parents.
Upkeep is set low ($u=2$), so survival is easy---the pressure is reproductive rather than lethal.

\subsection{LLM as agent policy}

Each agent's decision is produced by a single LLM call per turn.
The prompt provides full observability of the agent's local state but only partial observability of other agents (modulated by the SOC attribute).
No fine-tuning, few-shot examples, or behavioral instructions are provided; the LLM's pretrained policy determines all behavior.
This design tests whether the model's training data---encoding millennia of human strategic knowledge---is sufficient to produce adaptive behavior under environmental pressure.

\section{Experimental Setup}

\subsection{Experiment phases}

We conduct four phases of experiments totaling 22 completed runs:

\begin{itemize}[leftmargin=2em,itemsep=2pt]
  \item \textbf{Phase P1} (2 experiments): Validation runs for the v6.0 engine. Seed = 42. Confirmed agents make meaningful decisions with dual economy.
  \item \textbf{Phase P2} (13 experiments): Broad pressure sweep with upkeep $u \in \{0, 1, 2, 3, 4, 5, 6, 7, 8, 9, 10, 15\}$. Uses v6.0 engine with varying food/token node counts. Seed = 7. Identified pressure range of interest but confounded by simultaneously varying resource availability.
  \item \textbf{Phase P2b} (6 experiments): Controlled pressure sweep with upkeep $u \in \{2, 4, 5, 6, 7\}$. Uses v6.1 engine with constant resource nodes ($n_f = 8$, $n_t = 5$, $r_f = 3$, $r_t = 2$). Seed = 42. Two experiments were run at $u = 2$ (serving as replicates), with no experiment at $u = 3$ due to a configuration error.\footnote{EXP-020a was intended as upkeep=3 but inherited the phase default of upkeep=2, producing two independent runs at $u=2$ instead. We report both and note the gap at $u=3$.} This is our primary dataset for the Yerkes-Dodson analysis.
  \item \textbf{Phase V7} (1 experiment): Sexual selection baseline. Uses v7.0 engine with provider/chooser dynamics on a $7 \times 7$ grid. Upkeep $u = 2$ (low survival pressure). Seed = 42, 40 turns.
\end{itemize}

\subsection{Model and compute}

All experiments use Claude 3.5 Sonnet~\citep{anthropic2025claude} as the agent policy.
LLM calls are parallelized with a concurrency of 4.
Each experiment takes approximately 50--90 minutes on a single laptop, with approximately 18 seconds per LLM call.
Total compute for all 22 experiments: approximately 25 hours of wall-clock time and \$50--100 in API costs.

\subsection{Metrics}

We measure the following per experiment:

\begin{itemize}[leftmargin=2em,itemsep=2pt]
  \item \textbf{Trade count}: Number of successfully completed trades (our primary cooperation metric).
  \item \textbf{Attack count}: Total inter-agent attacks (aggression metric).
  \item \textbf{Survivors}: Agents alive at game end.
  \item \textbf{Game duration}: Number of turns before all-but-one agents die (or max turns reached).
  \item \textbf{Social action \%}: Proportion of actions that are TRADE or COMMUNICATE.
  \item \textbf{Shannon entropy}: Normalized entropy of the whole-game action distribution~\citep{shannon1948mathematical}:
  \[H_{norm} = -\frac{1}{\log k} \sum_{i=1}^{k} p_i \log p_i,\]
  where $k$ is the number of distinct action types.
\end{itemize}

\section{Results}

\subsection{Finding 1: Trade cooperation follows an inverted-U curve}

\begin{figure}[t]
  \centering
  \includegraphics[width=0.9\textwidth]{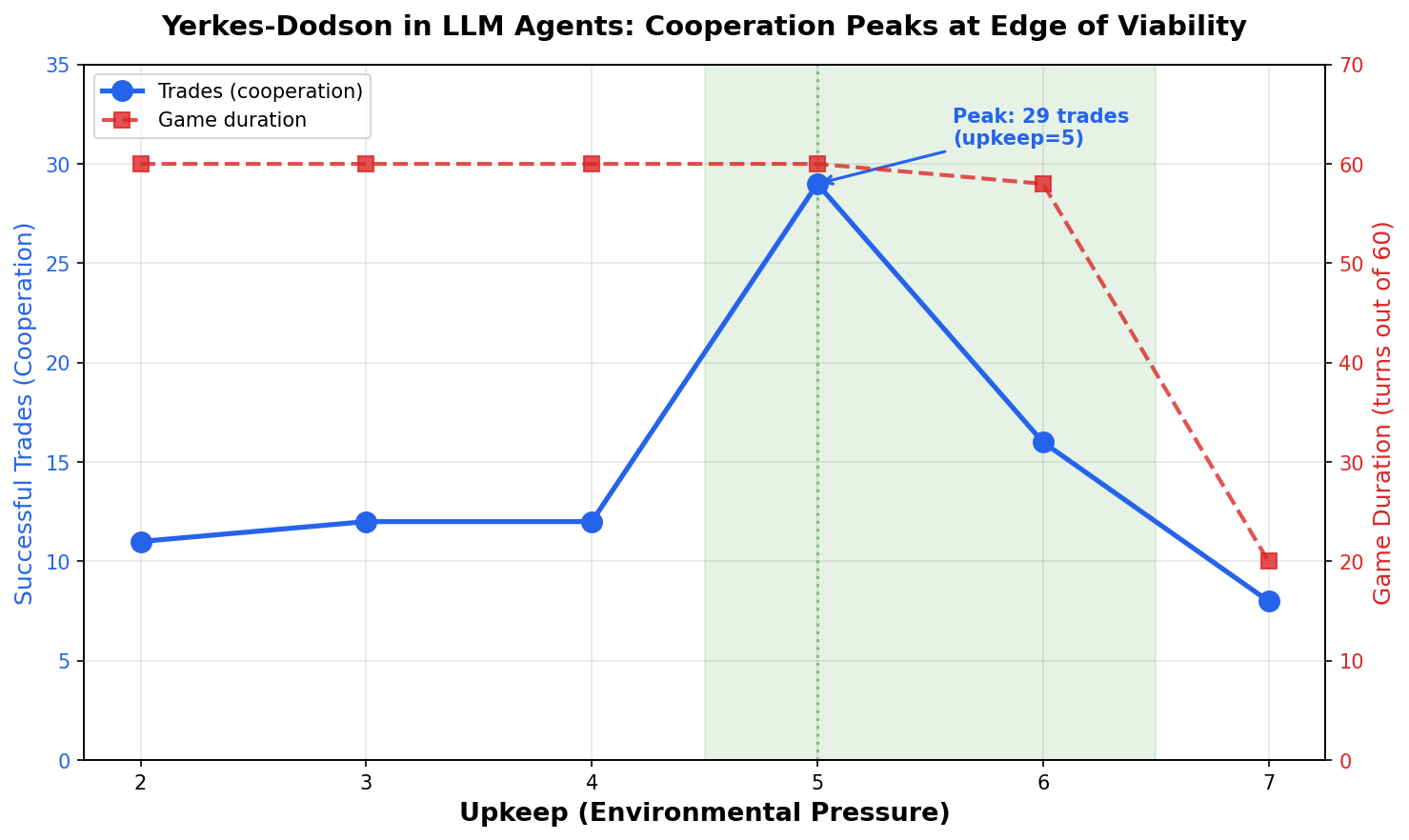}
  \caption{The Yerkes-Dodson curve for LLM agent cooperation. Left axis: successful trades (blue circles). Right axis: game duration in turns (red squares). The green shaded region marks the ``edge of viability'' where cooperation peaks. At upkeep = 7, the game collapses to 20 turns as agents die rapidly, eliminating opportunities for social interaction. Note: upkeep = 2 has two replicates (11 and 12 trades); the figure shows one representative point.}
  \label{fig:yerkes-dodson}
\end{figure}

Figure~\ref{fig:yerkes-dodson} shows our main result.
Across the P2b experiments, trade cooperation exhibits a clear inverted-U relationship with environmental pressure.
At low pressure (upkeep = 2), agents complete 11--12 trades over 60 turns (two replicates).
At medium pressure (upkeep = 5), trades spike to 29.
At high pressure (upkeep = 6--7), trades decline to 16 and 8, respectively, as agents increasingly die before they can cooperate.

This pattern closely mirrors the classical Yerkes-Dodson law: insufficient pressure produces behavioral stagnation (agents default to GATHER + MOVE loops), while excessive pressure collapses the behavioral repertoire to survival-only actions.
The ``sweet spot'' at upkeep = 5 creates conditions where agents face genuine scarcity---forcing them to seek cooperative solutions---without dying so quickly that cooperation cannot develop.

Table~\ref{tab:p2b-results} presents the full P2b results.
The social action percentage increases monotonically with pressure (2.8\% to 9.5\%), but this is confounded by the declining denominator (total actions decrease as agents die earlier).
The trade count, as an absolute measure, shows the inverted-U more clearly.

\begin{table}[t]
  \centering
  \caption{Phase P2b results. All experiments use v6.1 engine ($9 \times 9$ grid), 16 agents, constant resource nodes ($n_f=8$, $n_t=5$), seed = 42. Upkeep = 2 has two replicates (EXP-020b and EXP-020a). \textbf{Bold}: peak values.}
  \label{tab:p2b-results}
  \begin{tabular}{@{}lccccccc@{}}
    \toprule
    \textbf{Upkeep} & \textbf{Expt.} & \textbf{Trades} & \textbf{Attacks} & \textbf{Surv.} & \textbf{Dur.} & \textbf{Soc.\%} & \textbf{Entropy} \\
    \midrule
    2 & 020b & 11 & 76 & 3 & 60 & 2.8 & 0.764 \\
    2 & 020a & 12 & 85 & 4 & 60 & 4.4 & 0.787 \\
    4 & 020c & 12 & 63 & 2 & 60 & 3.9 & 0.791 \\
    \textbf{5} & \textbf{020d} & \textbf{29} & 61 & 2 & 60 & \textbf{8.4} & 0.864 \\
    6 & 020e & 16 & 39 & 1 & 58 & 6.5 & 0.861 \\
    7 & 020f & 8 & 19 & 1 & 20 & 9.5 & 0.892 \\
    \bottomrule
  \end{tabular}
\end{table}

\subsection{Finding 2: Extreme pressure causes behavioral collapse}

\begin{figure}[t]
  \centering
  \includegraphics[width=0.9\textwidth]{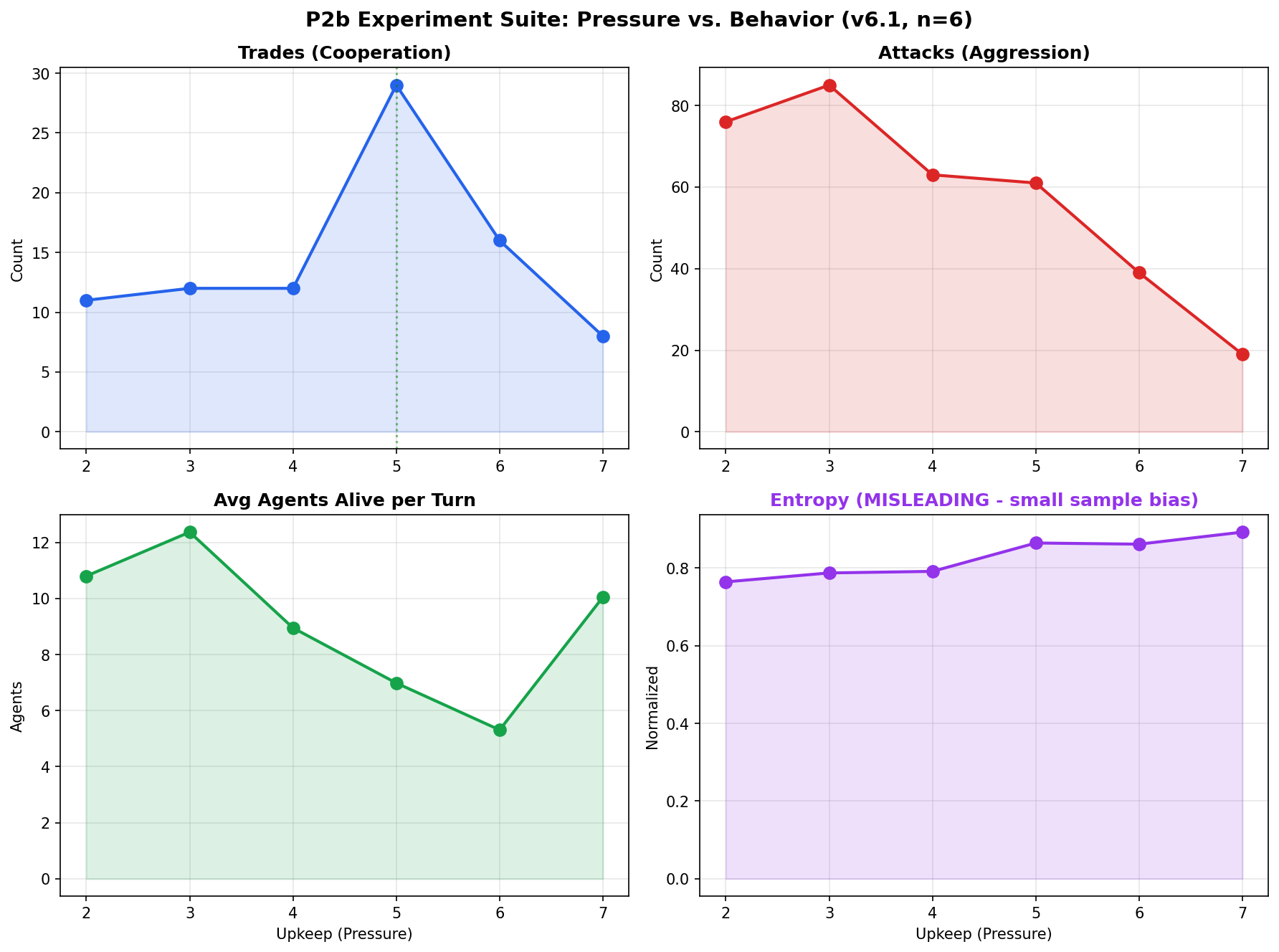}
  \caption{Multi-metric analysis of P2b experiments. Top-left: Trades show inverted-U (peak at upkeep = 5). Top-right: Attacks decrease monotonically with pressure as agents die before fighting. Bottom-left: Average agents alive per turn decreases with pressure. Bottom-right: Shannon entropy increases monotonically---a misleading artifact of small-sample bias (see Section~\ref{sec:entropy}).}
  \label{fig:multi-metric}
\end{figure}

Figure~\ref{fig:multi-metric} shows multi-metric analysis across pressure levels.
At high pressure (upkeep $\geq 7$), the game collapses rapidly.
In the P2 phase, upkeep levels of 8--15 reduced game duration to 5--12 turns, with agents converging on MOVE-dominated strategies (56--68\% of all actions).
Gathering success rates dropped below 60\% as competition for scarce nodes intensified.

The collapse follows a consistent pattern: (1) agents attempt to gather but nodes are insufficient, (2) MOVE percentage increases as agents search for resources, (3) social actions disappear entirely, (4) agents die in rapid succession.
At upkeep = 15 (``apocalypse'' pressure), the game lasted only 5 turns with 67.7\% MOVE actions and zero trades.

\subsection{Finding 3: Sexual selection eliminates aggression}

Table~\ref{tab:v7-results} compares the V7 sexual selection experiment with a representative P2b survival pressure run.
Under reproductive pressure, attacks drop to zero, while COMMUNICATE and REPRODUCE actions appear (absent in P2b).
The V7 experiment produced 3 offspring from 17 reproduction attempts, with 8 communication events---demonstrating that agents engage in costly signaling behavior when reproduction is at stake.

All 12 surviving agents (of 19 total, including 3 offspring) lived to the end of the 40-turn game, compared to the high mortality in P2b experiments.
The population grew from 16 to a peak of 18 before some agents died, illustrating that reproductive pressure maintains population size while survival pressure depletes it.

\begin{table}[t]
  \centering
  \caption{Sexual selection (V7) vs.\ survival pressure (P2b). V7 uses provider/chooser dynamics with low upkeep ($u=2$) on a $7 \times 7$ grid, 40 turns. P2b uses $9 \times 9$ grid, 60 turns.}
  \label{tab:v7-results}
  \small
  \begin{tabular}{@{}lcccccc@{}}
    \toprule
    \textbf{Expt.} & \textbf{Type} & \textbf{Agents} & \textbf{Surv.} & \textbf{Attacks} & \textbf{Trades} & \textbf{REPRO/COMM} \\
    \midrule
    V7-01a & Sexual sel. & 16$\to$19 & 12 & 0 & 6 & 17 / 8 \\
    020b (P2b) & Survival & 16 & 3 & 76 & 11 & --- / --- \\
    \bottomrule
  \end{tabular}
\end{table}

\subsection{Finding 4: Shannon entropy is misleading}
\label{sec:entropy}

A counterintuitive result is that Shannon entropy of the action distribution \emph{increases} monotonically with pressure (Table~\ref{tab:p2b-results}, last column: 0.764 $\to$ 0.892).
This appears to contradict the inverted-U hypothesis---but is an artifact of sample size.

At high pressure, agents die quickly, producing few total actions.
The survivors' actions are more uniformly distributed (fewer GATHER-dominated sequences), yielding higher entropy.
At upkeep = 7, only 201 total actions occur across 20 turns, compared to 648 at upkeep = 2 across 60 turns.
The entropy increase reflects the breakdown of the GATHER-dominated behavioral mode, not genuine behavioral complexity.

This finding has methodological implications: whole-game Shannon entropy is not a reliable proxy for behavioral complexity in environments where pressure affects population survival.
Per-turn entropy or action diversity metrics that control for population size would be more appropriate.

\section{Discussion}

\subsection{The sweet spot: edge of viability}

Our results suggest that the optimal environmental pressure for LLM agent cooperation lies at the ``edge of viability''---the pressure level where agents face genuine existential threat but have sufficient time to develop cooperative strategies.
At upkeep = 5, agents can survive approximately 12 turns on initial resources alone, compared to 30+ turns at upkeep = 2.
This creates a window of urgency that promotes trade without immediately killing agents.

The analogy to the Yerkes-Dodson law is direct: low arousal (low upkeep) produces behavioral stagnation, high arousal (high upkeep) produces performance collapse, and intermediate arousal produces peak performance.
The key insight for AI agent curriculum design is that environmental pressure can serve as a training curriculum even without gradient-based optimization---the LLM's pretrained policy adapts behaviorally to the pressure level.

We note a limitation: the gap at upkeep = 3 (due to a configuration error producing two runs at upkeep = 2 instead) means our curve has uneven spacing.
However, the two replicates at upkeep = 2 provide a consistency check: they produced 11 and 12 trades respectively, suggesting low inter-run variance at this pressure level.

\subsection{Sexual selection as ``softer'' pressure}

Survival pressure has an inherent limitation: it kills subjects, reducing the population available for observation and interaction.
Sexual selection offers an alternative: all agents survive, but not all reproduce.
This creates competitive pressure---agents must signal quality, communicate, and position themselves strategically---without the population collapse that limits high-pressure survival experiments.

The complete elimination of attacks under sexual selection is striking.
Under survival pressure (P2b), attacks constitute 9--14\% of actions even at moderate pressure.
Under sexual selection, agents apparently ``decide'' that attacking potential mates or their offspring is counterproductive.
This aligns with biological theory: sexual selection drives the evolution of display, communication, and courtship rather than combat~\citep{darwin1871descent, zahavi1975mate}.

\subsection{LLM pretrained policy as genetic memory}

A remarkable feature of our setup is that agents receive no behavioral instructions, few-shot examples, or fine-tuning.
All observed behavior---cooperation, aggression, resource hoarding, strategic reproduction---emerges from the LLM's pretrained policy when confronted with the environmental prompt.
This is analogous to biological organisms whose behavior is shaped by evolutionary history (encoded in DNA) rather than individual learning.

The LLM's training data effectively serves as ``genetic memory''---encoding millennia of human strategic knowledge that agents can draw upon.
This has implications for agent design: rather than constraining behavior through hard-coded rules, environmental prompts that activate appropriate reasoning chains may be more effective.

\section{Limitations}

Our study has several important limitations:

\begin{enumerate}[leftmargin=2em,itemsep=2pt]
  \item \textbf{Single run per configuration.} Most experiments represent a single game with one random seed. We cannot compute confidence intervals or test statistical significance. The upkeep = 2 level has two replicates showing consistent results (11 vs.\ 12 trades), but other levels have $n = 1$.
  \item \textbf{Single LLM model.} All experiments use Claude 3.5 Sonnet. Different models exhibit distinct behavioral profiles~\citep{fan2025understanding, chen2025strategic}; the stress-performance curve likely differs across models.
  \item \textbf{Small scale.} 16 agents on a $9 \times 9$ grid over 40--60 turns is far from the scale of real multi-agent systems. Scaling effects on the Yerkes-Dodson curve are unknown.
  \item \textbf{Missing pressure level.} Due to a configuration error, the P2b dataset has no experiment at upkeep = 3, leaving a gap in the curve between upkeep = 2 and 4.
  \item \textbf{Entropy metric limitations.} As discussed in Section~\ref{sec:entropy}, our complexity metric (Shannon entropy) proved confounded. Per-turn metrics were not computed in these experiments.
  \item \textbf{No cross-generation learning.} Each game starts fresh; agents do not inherit strategies from previous games. The stress-performance relationship may differ with inter-game memory.
\end{enumerate}

\section{Future Work}

Several directions follow from our results:

\begin{itemize}[leftmargin=2em,itemsep=2pt]
  \item \textbf{Multi-model arena (V8):} Different LLMs (Claude, GPT-4o, DeepSeek-R1, Gemini, Llama) as different ``species'' in the same arena, with each model's pretrained behavioral profile serving as its genotype. This would test whether stress curves vary by model and whether behavioral diversity produces richer dynamics.
  \item \textbf{Statistical rigor:} Multiple seeds per configuration to compute confidence intervals and perform significance testing on the inverted-U pattern. Fill the gap at upkeep = 3.
  \item \textbf{Per-turn metrics:} Behavioral complexity measured at each turn rather than over the whole game, controlling for population size effects.
  \item \textbf{Cross-generation memory:} Strategy inheritance between games, enabling evolutionary dynamics where successful strategies propagate through reproduction.
  \item \textbf{Stress curve as product:} A systematic characterization of optimal pressure levels for different LLM models could serve as a curriculum design tool for AI agent development.
\end{itemize}

We are actively pursuing the multi-model arena (V8) and multi-seed validation as immediate next steps. Code, experiment data, and analysis scripts are available at \url{https://github.com/anthroos/survival-arena}.

\section{Conclusion}

We have presented the first empirical evidence that LLM multi-agent systems exhibit a Yerkes-Dodson stress-performance curve.
Across 22 experiments in a grid-world survival arena, we find that cooperative behavior peaks at intermediate environmental pressure and declines under both low and extreme conditions.
Trade cooperation reaches 29 at the optimal pressure level (upkeep = 5) compared to 11--12 at low-pressure baselines, while extreme pressure eliminates social behavior within 5--12 turns.

Our results also introduce sexual selection as a promising alternative pressure mechanism for LLM agent environments: it creates competitive dynamics without lethal consequences, eliminates aggression, and produces communicative behavior absent under survival pressure.

These findings have practical implications for AI agent curriculum design.
Rather than optimizing model weights through gradient descent, environmental pressure can serve as a training curriculum that shapes agent behavior through the LLM's pretrained policy.
Finding the optimal pressure---the AI Yerkes-Dodson curve---for each model and task domain may be a key design parameter for the next generation of multi-agent AI systems.

\bibliographystyle{plainnat}
\bibliography{references}

\newpage
\appendix

\section{Full Experiment Results}
\label{app:full-results}

Table~\ref{tab:all-experiments} shows all 22 completed experiments across four phases.

\begin{table}[h]
  \centering
  \caption{Complete experiment results across all phases. P2 varies both upkeep and node counts (confounded). P2b varies only upkeep with constant nodes. V7 uses sexual selection. $^\dagger$EXP-020a ran at upkeep=2 (not 3 as originally intended).}
  \label{tab:all-experiments}
  \small
  \begin{tabular}{@{}llccccccc@{}}
    \toprule
    \textbf{ID} & \textbf{Phase} & \textbf{Upkeep} & \textbf{Nodes} & \textbf{Trades} & \textbf{Attacks} & \textbf{Surv.} & \textbf{Turns} & \textbf{Entropy} \\
    \midrule
    EXP-010a & P1 & 2 & 8/5 & 37 & 58 & 9 & 50 & 0.883 \\
    EXP-010b & P1 & 2 & 8/5 & 4 & 10 & 7 & 60 & 0.488 \\
    \midrule
    EXP-011a & P2 & 0 & 8/5 & 0 & 6 & 7 & 60 & 0.221 \\
    EXP-011b & P2 & 1 & 8/5 & 3 & 16 & 8 & 60 & 0.430 \\
    EXP-011c & P2 & 2 & 8/5 & 0 & 22 & 7 & 60 & 0.525 \\
    EXP-011d & P2 & 2 & 6/4 & 2 & 33 & 8 & 60 & 0.600 \\
    EXP-011e & P2 & 3 & 6/3 & 5 & 35 & 6 & 60 & 0.667 \\
    EXP-011f & P2 & 4 & 4/2 & 6 & 10 & 3 & 60 & 0.652 \\
    EXP-011g & P2 & 5 & 3/1 & 3 & 5 & 1 & 60 & 0.682 \\
    EXP-011h & P2 & 7 & 2/1 & 0 & 2 & 2 & 12 & 0.633 \\
    EXP-011i & P2 & 10 & 1/1 & 0 & 1 & 1 & 8 & 0.532 \\
    EXP-011j & P2 & 15 & 1/1 & 1 & 4 & 3 & 5 & 0.626 \\
    EXP-011k & P2 & 6 & 3/1 & 0 & 2 & 1 & 16 & 0.542 \\
    EXP-011l & P2 & 8 & 2/1 & 2 & 5 & 2 & 10 & 0.619 \\
    EXP-011m & P2 & 9 & 1/1 & 1 & 5 & 2 & 9 & 0.685 \\
    \midrule
    EXP-020a$^\dagger$ & P2b & 2 & 8/5 & 12 & 85 & 4 & 60 & 0.787 \\
    EXP-020b & P2b & 2 & 8/5 & 11 & 76 & 3 & 60 & 0.764 \\
    EXP-020c & P2b & 4 & 8/5 & 12 & 63 & 2 & 60 & 0.791 \\
    EXP-020d & P2b & 5 & 8/5 & \textbf{29} & 61 & 2 & 60 & 0.864 \\
    EXP-020e & P2b & 6 & 8/5 & 16 & 39 & 1 & 58 & 0.861 \\
    EXP-020f & P2b & 7 & 8/5 & 8 & 19 & 1 & 20 & 0.892 \\
    \midrule
    EXP-V7-01a & V7 & 2 & 8/5 & 6 & 0 & 12 & 40 & 0.689 \\
    \bottomrule
  \end{tabular}
\end{table}

\section{Action Distribution by Pressure Level}
\label{app:action-dist}

Table~\ref{tab:action-dist} shows the percentage breakdown of actions at each P2b pressure level.

\begin{table}[h]
  \centering
  \caption{Action distribution (\%) for P2b experiments. u=2 shows EXP-020b; EXP-020a (also u=2) has similar distribution (43.1/29.9/11.5/8.8/4.4/2.3\%).}
  \label{tab:action-dist}
  \begin{tabular}{@{}lccccc@{}}
    \toprule
    \textbf{Action} & \textbf{u=2} & \textbf{u=4} & \textbf{u=5} & \textbf{u=6} & \textbf{u=7} \\
    \midrule
    GATHER  & 44.6 & 45.6 & 40.6 & 42.2 & 38.8 \\
    MOVE    & 29.5 & 26.6 & 22.7 & 22.7 & 23.4 \\
    ATTACK  & 11.7 & 11.7 & 14.6 & 12.7 &  9.5 \\
    TRAIN   &  9.6 &  8.6 &  9.8 &  7.8 & 10.0 \\
    TRADE   &  2.8 &  3.9 &  8.4 &  6.5 &  9.5 \\
    REST    &  1.9 &  3.5 &  4.1 &  8.1 &  9.0 \\
    \midrule
    \textbf{Total actions} & 648 & 537 & 419 & 308 & 201 \\
    \bottomrule
  \end{tabular}
\end{table}

\section{Agent Prompt Template}
\label{app:prompt}

The following is the core structure of the prompt provided to each agent at each turn (v6.1 engine). Specific values are filled dynamically.

\begin{verbatim}
You are Agent [ID] in a survival arena.

YOUR STATUS:
- Position: ([x], [y])
- Food: [food], Tokens: [tokens], Health: [hp]/[max_hp]
- Attributes: STR=[s] SPD=[sp] INT=[i] SOC=[so] END=[e] CHA=[c]

NEARBY AGENTS (within 2 cells):
[List of visible agents with position, approximate resources]

AVAILABLE ACTIONS:
GATHER, MOVE [direction], ATTACK [target_id],
TRADE [target_id] [offer] [request], REST, TRAIN [attribute]

Choose exactly ONE action. Respond with the action name
and parameters only.
\end{verbatim}

Note: No behavioral hints, cooperation encouragements, or survival tips are provided. The LLM's pretrained policy determines all strategic reasoning.

\section{V7 Sexual Selection Mechanics}
\label{app:v7-mechanics}

The V7 engine introduces reproductive pressure through asymmetric costs and mate choice:

\begin{enumerate}[leftmargin=2em,itemsep=2pt]
  \item \textbf{Provider proposes:} A provider agent at the same position as a chooser can select REPRODUCE, specifying the target chooser. Cost: 6 food + 3 tokens.
  \item \textbf{Chooser evaluates:} The chooser receives a separate LLM call with the provider's visible stats (attributes, resources, vitality). The chooser decides accept/reject.
  \item \textbf{Offspring creation:} If accepted, an offspring is created at the same position with attributes averaged from both parents plus Gaussian noise ($\sigma = 1$, clipped to $[1, 10]$). Offspring vitality is the mean of parents' vitality.
  \item \textbf{Chooser cost:} The chooser pays 12 food + 5 tokens regardless of the decision to reproduce. This reflects the asymmetric parental investment~\citep{trivers1972parental}.
\end{enumerate}

The vitality system creates a quality signal: radiant agents (vitality 9--10) start with more resources and produce higher-vitality offspring, but the cost of reproduction is the same regardless of vitality.
This incentivizes choosers to prefer high-vitality providers, creating selection pressure for resource accumulation as a quality signal---analogous to the handicap principle~\citep{zahavi1975mate}.

\end{document}